\documentclass[conference]{IEEEtran}
\usepackage{tabularx}
\usepackage{booktabs}
\usepackage[hidelinks]{hyperref}
\usepackage{url}
\usepackage{xurl}
\IEEEoverridecommandlockouts
\usepackage{cite}
\usepackage{amsmath,amssymb,amsfonts}
\usepackage{algorithmic}
\usepackage{graphicx}
\usepackage{textcomp}
\usepackage{xcolor}
\def\BibTeX{{\rm B\kern-.05em{\sc i\kern-.025em b}\kern-.08em
    T\kern-.1667em\lower.7ex\hbox{E}\kern-.125emX}}
\begin{document}

\title{\vspace{1em}\Large{When Stopping Fails: Rethinking Minimal Risk Conditions through Human-Interactive Autonomous Driving for Safe Transportation Systems
}}

\author{
\IEEEauthorblockN{Yash Tandon}
\IEEEauthorblockA{
\textit{UC San Diego}}

\and
\IEEEauthorblockN{Giovanni Tapia Lopez}
\IEEEauthorblockA{
\textit{UC Merced}}

\and
\IEEEauthorblockN{Marcus Blennemann}
\IEEEauthorblockA{
\textit{UC San Diego}}

\and
\IEEEauthorblockN{Mohan Trivedi}
\IEEEauthorblockA{
\textit{UC San Diego}}

\and
\IEEEauthorblockN{Ross Greer}
\IEEEauthorblockA{
\textit{UC Merced}}
}

\maketitle

\begin{abstract}
Autonomous vehicles (AVs) are increasingly deployed in urban environments, yet their safety frameworks remain primarily designed around collision avoidance and minimal risk condition (MRC) behaviors such as slowing or stopping when uncertainty arises. Although effective in reducing immediate crash risk, real-world deployments indicate that stopping alone does not guarantee safe integration into human-governed roadway systems. Incidents reported by municipalities and public records show that AV fallback behaviors can obstruct traffic, interfere with emergency response operations, and create accessibility challenges for passengers and pedestrians. This paper presents an analysis of publicly documented incidents involving AV stopping behavior and human–AV interaction failures. We categorize these incidents according to limitations in perception, planning, and control within current AV architectures. Using this taxonomy, we identify key gaps in existing safety paradigms, particularly the lack of mechanisms for interpreting human authority, responding to multimodal instructions, and adapting to dynamic, socially regulated traffic conditions. We then review emerging research directions that support human-interactive perception, language-grounded and accessibility-aware planning, and assisted control through remote guidance and teleoperation. The analysis highlights the need to augment current AV safety frameworks with capabilities that enable cooperative interaction with human agents and infrastructure. These findings suggest that reliable urban deployment of AVs requires moving beyond passive fallback strategies toward human-interactive autonomy.
\end{abstract}


\section{Introduction}

In recent years, autonomous vehicles (AVs) have been on the rise and are now commercially deployed in multiple urban environments through ride-hailing services \cite{lee_2024_waymos}. Deployment at such a scale raises important questions about safety and security \cite{partnersforautomatedvehicleeducationpave_2024_pave} \cite{dagostino_2024_experiences}. Not only do we require AVs to avoid collisions, but also integrate safely into human-governed infrastructure systems, just as one would expect a human driver to. This includes handling adverse scenarios such as: emergency response scenes, temporary road closures, crowd control situations, informal human traffic direction, and accessibility-related pickups/drop-offs \cite{dagostino_2024_experiences} \cite{zhou_2023_sf} \cite{nicholson_2023_cpuc} \cite{tumlin_2023_san} \cite{californiapublicutilitiescommission_2023_re}. These scenarios expose a fundamental distinction between laboratory safety validation and real-world cooperative integration \cite{white_2023_san}. As we move forward with the development of AV safety, the challenge is no longer solely perception accuracy or collision avoidance, but cooperative participation in human-controlled roadway ecosystems.

\begin{figure}
    \centering
    \includegraphics[width=.95\linewidth]{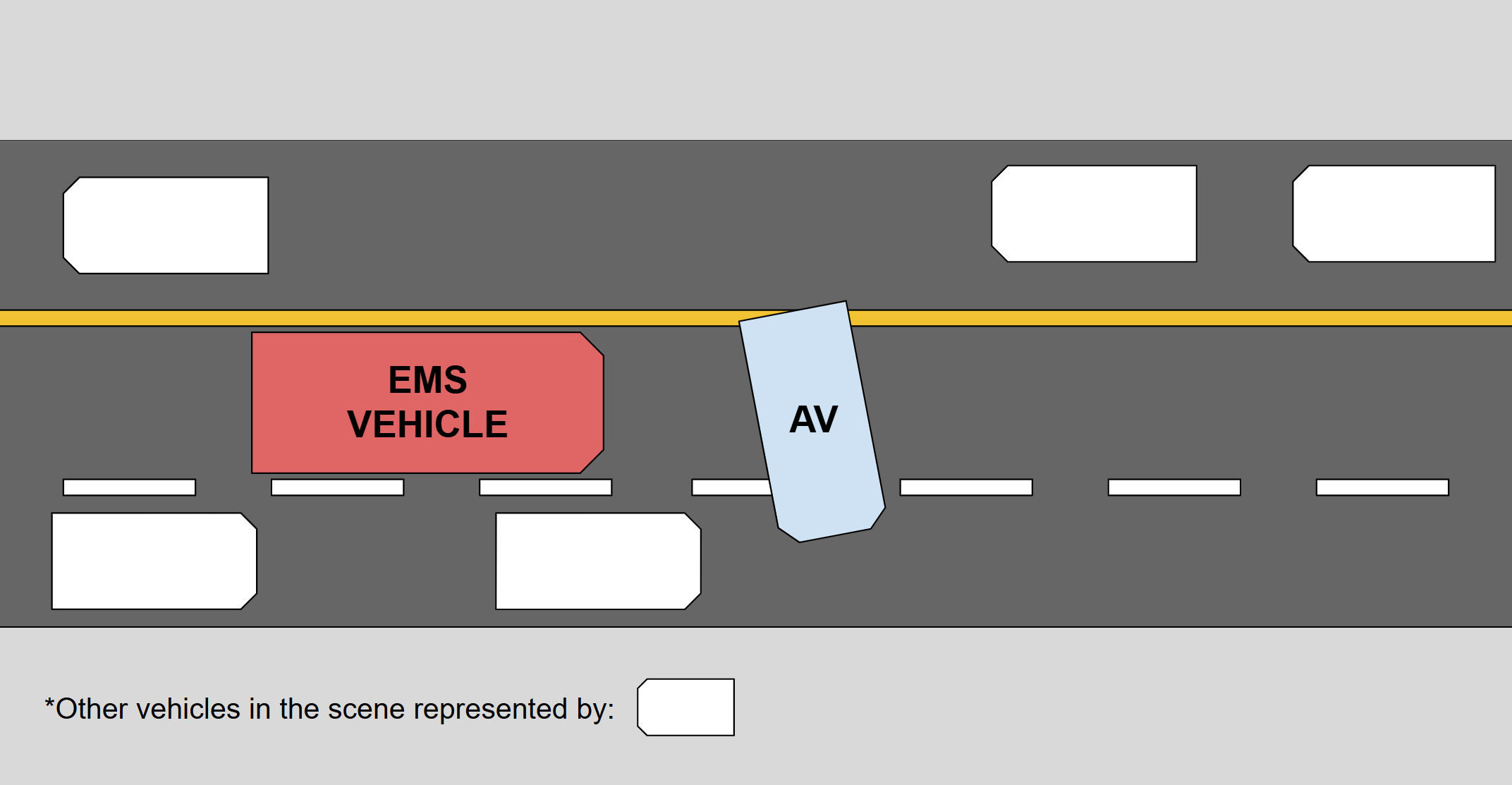}
    \caption{Illustrative reconstruction of a reported incident in Austin, Texas, where a commercially-deployed AV unexpectedly stopped near an active emergency scene, obstructing an EMS until first responders intervened and manually steered away the vehicle.}
    \label{fig:austin_incident}
\end{figure}

Currently, AVs follow a safety framework relying on the concept of minimal risk condition, which instructs the AV to slow down, pull over, stop, and turn on the hazard lights, whenever uncertainty arises \cite{onroadautomateddrivingoradcommittee_2021_j3016_202104}. In this case, the “safe stop” philosophy ensures that the AV is taking utmost precaution to avoid collisions and system failures. However, the minimal risk condition is designed to reduce uncertainty, but it does not guarantee non-obstruction, non-interference, or cooperative compliance \cite{automateddrivingbehaviorsconsortium_2019_minimal}. Real-world deployments suggest that stopping can cause obstruction, delay emergency responses, and transfer additional, unnecessary workload to human responders \cite{ding_2025_waymos} \cite{lu_2023_lost}. For example, after a fatal shooting in Austin, Texas, a commercially-deployed AV was reported to have stalled in the middle of roadway obstructing an Emergency Medical Services (EMS) vehicle responding to the scene; the shooting led to 3 deaths and 14 hospitalizations, and EMS services to such situations are extremely vital and time-sensitive. We depict a simplification of the incident in Figure 1 to illustrate the AV placement \cite{cobler_2026_waymo}. Although the vehicle was not involved in a collision, the fallback stopping behavior interfered with a time-critical emergency situation. Similarly, municipal agencies, such as the San Francisco Municipal Transportation Agency (SFMTA), have voiced their concerns by reporting recurring incidents involving AVs blocking fire stations, intruding emergency scenes, behaving unpredictably near responders, and requiring complex intervention before manual takeover of the vehicle \cite{nicholson_2023_cpuc} \cite{californiastateassembly_2024_assembly}. In complex, time-critical environments, stopping is not equivalent to safety.

Beyond scenarios involving emergency responders, human-interaction gaps in current AVs also affect individuals with disabilities relying on AVs for mobility, pedestrians in non-standard traffic scenarios, and traffic management during road constructions. To ensure maximum accessibility, AVs must be capable of reliable curb approach, context-aware pickup/dropoff, and multimodal communication \cite{dagostino_2024_experiences} \cite{mkaufman_2022_autonomous} \cite{yousfi_2025_automated}. In addition, perception must be inclusive to detect passengers with disabilities, which is often underrepresented in training data and simulations \cite{dredf_2023_autonomous}. These cases reinforce that safety is not solely a function of collision avoidance, but of responsiveness to human needs and context-dependent authority.

The AV stack is constantly improving in its core functionality in object detection, trajectory prediction, and collision-free planning thanks to ongoing advancements in machine learning and suitable hardware to handle intensive computational decision-making \cite{lin_2018_the, nebot2026eraendtoendautonomytransitioning, hu_2022_planningoriented, greer2025language}. However, to ensure safe societal integration, recent incidents suggest that planning algorithms must go beyond the core object perception and surround agent motion prediction capabilities of AVs; in addition to scene understanding, it is essential that AVs recognize authority, interpret instructional commands through voice and gestures, and cooperate through effective bidirectional communication. Safe integration therefore requires AV systems to be designed not only for obstacle avoidance, but for authority-aware, socially legible, and cooperative interaction with human agents, which requires human-interactive perception, planning, and control \cite{mkaufman_2022_autonomous} \cite{advocatesforhighwayandautosafety_2020_autonomous} \cite{cano_2024_exclusive}.

This survey paper makes several contributions toward understanding the limitations of autonomous vehicles deployed in urban, human-governed roadways. First, we objectively summarize publicly documented incidents involving AV-human interaction across first responder and accessibility contexts. Then, we identify possible reasons and technological gaps related to for the observed failures with respect to the current AV safety paradigms. Building on this, we then demonstrate the need for human-interactive perception, planning, control, and communication layers integrated into AVs. Finally, we review emerging technological approaches that could enable robust, responsive, human-interactive AV behavior.

\section{Background and Safety Framing}
\subsection{AV Safety Architecture}
The decision-making pipelines powering AV systems have traditionally followed a modular perception-planning-control structure, although recent end-to-end and reasoning-capable approaches are beginning to challenge this paradigm. \cite{nebot2026eraendtoendautonomytransitioning}. AVs consist of layers of perception, prediction, planning, and control modules. The pipeline begins with perception modules processing data from multimodal sensory inputs such as cameras, lidar, and radar to detect, recognize, and segment objects in the surroundings. Prediction modules then estimate trajectories of the surrounding objects, and consequently, planning modules compute the ego vehicle trajectory to avoid collision whilst obeying traffic laws. Lastly, the control system executes the planned motion through powering the mechanical components to enable appropriate steering, acceleration, and braking \cite{lin_2018_the} \cite{hu_2022_planningoriented}.

The modular framework primarily frames safety as collision avoidance, particularly when uncertainty arises \cite{hu_2022_planningoriented}. Uncertainty is evaluated through confidence measures in object detection, trajectory forecasting, and motion planning \cite{arnez_2020_a}. The main strategy for the AV in this architecture is the concept of a minimal risk condition (MRC). MRC, defined in the SAE J3016, is the condition that instructs the autonomous system to trigger a fallback behavior whenever the system determines it cannot safely function within its operational design domain (ODD) \cite{onroadautomateddrivingoradcommittee_2021_j3016_202104}. Typical fallback behaviors include slowing down, pulling over to the curb, stopping, and activating hazard signals \cite{automateddrivingbehaviorsconsortium_2019_minimal}.

MRC assumes that reducing motion of the vehicle also reduces the risk of accidents. This logic aligns with traditional safety principles because minimizing dynamic interaction with the surroundings will eliminate the possibility of colliding or performing unsafe maneuvers \cite{onroadautomateddrivingoradcommittee_2021_j3016_202104} \cite{automateddrivingbehaviorsconsortium_2019_minimal}. For instance, if the AV detects unusual uncertainty due to mechanical failure or sensor degradation caused by adverse weather conditions, it’s best that the AV stops as soon as possible and waits until the issue is resolved. 

However, this behavior assumes that the surrounding environment will be able to adapt to a stationary or slow moving vehicle. AV may cause obstruction, delay, or confusion, particularly in dynamic, authority-driven environments \cite{nicholson_2023_cpuc} \cite{saeinternational_2024_195}. Integration of authority recognition, cooperative negotiation, or real-time compliance with human language commands is still an ongoing problem in the field of AVs \cite{roy_2024_doscenes} \cite{chang_2024_it}.

Another emerging effort relevant to autonomous vehicle interaction with humans is the IEEE P3474, a draft standard that focuses on aligning autonomous system behavior with human expectations and intentions. Rather than defining operational states or fallback procedures, the standard proposes representative interaction scenarios and measurable evaluation metrics to assess whether an autonomous agent behaves in ways that are understandable and predictable to human road users. In the context of autonomous driving, this includes situations involving ambiguous right-of-way decisions, pedestrian interactions, or other social driving behaviors where human drivers rely on shared norms. As a result, IEEE P3474 complements frameworks such as SAE J3016 by shifting attention from defining safe operating conditions to evaluating whether the vehicle’s decisions are consistent with how humans expect vehicles to behave in mixed traffic environments\cite{agent2025p3474}.

\subsection{AV Interaction with Human Authority} 
Static infrastructure, such as traffic lights, signs, and lane markings, are not the only components governing our roads. Under certain circumstances, human authority takes precedence in regulating the behavior of traffic. In particular, first responders, work zone workers, and traffic guards, are authorized to take control of the road \cite{nicholson_2023_cpuc} \cite{tumlin_2023_san} \cite{californiastateassembly_2024_assembly} \cite{saeinternational_2024_195}. For instance, police officers may overrule traffic signals, halt vehicles mid-intersection, or use hand-signals to manually direct vehicles through another, nonstandard lane. Consequently, the traffic rules that must be followed in such scenarios become dynamic and authority-driven, and no longer strictly defined by the standard traffic code.

Human drivers are socially conditioned to understand when and how to follow authoritative signals by recognizing features like uniforms, hand gestures, flashing lights, barricades, and custom traffic signs. Through eye contact, implicit negotiation, and contextual reasoning in real-time, human drivers are able to adapt their behavior to the established temporary rules \cite{chang_2024_it, rasouli_2022_autonomous}.

However, AV systems are trained to operate under defined, rule-based traffic conditions \cite{lin_2018_the}, and rare scenarios are not well represented in the data \cite{nebot2026eraendtoendautonomytransitioning, shriram2025towards, chen2025robust}. Effective compliance in authority-driven scenarios requires the AV to reinterpret the operational context and, as a result, revise its motion planning to respond directly to human instructions. These capabilities are not explicitly modeled in the perception-planning-control abstractions of AV systems \cite{lin_2018_the} \cite{hu_2022_planningoriented}.

Additionally, first responders operate in environments that have seconds-scale decision cycles, and thus delays to their work can result in detrimental consequences for surrounding individuals \cite{nicholson_2023_cpuc} \cite{saeinternational_2024_195}. Current AVs lack robust interactive capabilities, and overriding their decisions can only be achieved through remote assistance or manual escalation pathways. Regulatory proceedings have put across that this introduces latency and burdens first responders with additional coordination steps \cite{nicholson_2023_cpuc} \cite{tumlin_2023_san}. Authority-driven roadways demand immediate, unambiguous compliance from surrounding vehicles to ensure safety; however, such interaction models are fundamentally misaligned with these requirements.

\subsection{Accessibility Capabilities in AVs}
Accessibility considerations in AVs is an important research focus, especially as AVs expand into dense urban environments, where user requirements includes the needs of users of wheelchairs or other mobility devices \cite{dredf_2023_autonomous} \cite{yousfi_2025_automated}. Prior work in this field includes improving assistive device (e.g. wheelchairs) recognition \cite{dvilasobern_2025_a} and more recently, assessment and detection of accessibility-relevant infrastructure on the roads such as curb ramps and crosswalk accessibility features through detailed geospatial datasets \cite{martnezchao_2024_urban} \cite{omeara_2022_rampnet}. In the context of AVs, these contributions reduce collisions and improve route planning that account for mobility constraints (i.e. choosing feasible pickup and drop-off (PUDO) locations for passengers) \cite{mkaufman_2022_autonomous}.

However, static detection alone does not enable accessibility in AVs. Accessibility also requires meaningful interaction with passengers who may require assistance, clarification, or adaptive, multimodal communication \cite{dredf_2023_autonomous} \cite{lin_2018_the}. This includes everything from a passenger signaling readiness from an unexpected location to a user providing verbal instructions to reposition the vehicle due to safety concerns \cite{chang_2024_it}. In addition, the multimodal communication should be bidirectional to enable responding to a diverse range of passenger needs. Passengers with visual impairments may require aural or haptic cues to find and enter the AV. Likewise, passengers with cognitive disabilities may aid from simplified instructions \cite{dredf_2023_autonomous}. Accessibility is often framed as a routing or localization problem for AVs, but in reality, accessible autonomy should be an iterative and interactive process \cite{mkaufman_2022_autonomous}. For example, an AV may be capable of finding and parking next to a wheelchair accessible curb cut with accessible routing; however, it may fail to recognize that there is construction equipment blocking the ramp, making it unsuitable for the passenger, requiring human interaction or instruction to create accessible and safe use of the transportation system.

\section{Incident Analysis and Potential Improvements}

\begin{table*}
    \centering
    \caption{AV Stopping Incidents}
    \label{AV Incident Types}
    \begin{tabularx}{\textwidth}{l c l X X}
        \toprule
        Source & Year & Failure type & Incident Summary & Areas for improvement \\
        \midrule
        KTVU News\cite{kafton_2022_video} & 2022 & Control & An AV was pulled over by San Francisco police because it was driving at night without headlights on. When stopped, it briefly moved forward after the officer walked away before stopping again with hazard lights on & Improve environmental safety checks (specifically, ensure headlights are on at night). Improve human interactivity so vehicle can understand natural language instruction \\
        \midrule
        KTLA 5 news\cite{dallow_2025_robot} & 2025 & Perception \& Planning & A food delivery vehicle blocked the way of a fire truck & Better scene interpretation, reasoning, and decision-making \\
        \midrule
        CalMatters hearing\cite{californiastateassembly_2024_assembly}& 2024 & Control \& Planning & Driverless cars ``stop suddenly,'' impeding traffic and causing accidents. & Improve motion planning and traffic flow logic to reduce abrupt stops. \\
        \midrule
        CalMatters hearing\cite{californiastateassembly_2024_assembly}& 2024 & Control \& Planning & AV blocked emergency vehicles, preventing police response to a mass shooting. & Better emergency-vehicle detection and cooperative behavior. \\
        \midrule
        CalMatters hearing\cite{californiastateassembly_2024_assembly}& 2024 & Perception & AVs ``drove through emergency scenes and into downed wires.'' & The AV's ability to recognize rarer scenarios needs to be improved \\
        \midrule
        CalMatters hearing\cite{californiastateassembly_2024_assembly}& 2024 & Perception \& Control & One AV ``drove away from police officers during a vehicle stop.'' & AV's ability to interpret natural language instructions \\
        \midrule
        CalMatters hearing\cite{californiastateassembly_2024_assembly}& 2024 & Perception \& Control & A pedestrian was dragged under an autonomous vehicle after being hit by a human driver; initially not fully disclosed. & Stricter reporting and review of crash footage; improve obstacle response behavior. \\
        \midrule
        SFMTA\cite{nicholson_2023_cpuc}& 2023 & Control \& Planning & Commercially-operated robotaxis have repeatedly made unplanned stops in travel lanes, blocking vehicle traffic, transit buses, light rail, and streetcars & AVs must be able to respond to more complex scenarios without simply stopping \\
        \midrule
        KXAN \& Austin Police\cite{kxan} & 2024 & Planning & During severe weather in Austin, several commerically deployed AVs pulled over and stopped in locations that partially blocked traffic, including a ramp area. The vehicles stopped as part of their safety response to bad weather, but their stopping positions obstructed traffic. & Improve fallback stopping strategy so autonomous vehicles select safe pull-over locations that do not block lanes, ramps, or critical traffic areas during weather-related service pauses. \\
        \midrule
        California Court Case\cite{tumlin_2023_san} & 2023 & Control & Around 10 robotaxis suddenly stopped operating at the same time in the North Beach neighborhood of San Francisco. The vehicles stopped in travel lanes, clustered together, and could not move or clear the roadway. This caused traffic congestion and blocked street access in a neighborhood with narrow streets. & Improve failure control systems so downed cars pull over to a safe area before stopping\\
        \midrule
        Axios Austin\cite{cobler_2026_waymo} & 2026 & Perception \& Planning &  A commercially-deployed AV reportedly stopped in roadway near active fatal shooting scene, obstructing EMS vehicle responding to the emergency & Authority-aware yielding, emergency corridor planning, responder communication interface\\
    \bottomrule
    \end{tabularx}
\end{table*}

As more AVs are being deployed publicly, there is an increased concern about whether they may cause problems in their urban environment. These concerns raise the question of which failure modes occur most frequently and how they may be corrected. An important factor contributing to these concerns lies in how AVs interpret risk and safety within their MRC framework, which assesses the best time for bringing the vehicle to a stop. Though this approach allows AVs to avoid some incidents, it can lead to others where inappropriate stopping causes a problem.  These new issues can take many forms, such as interfering with first responders to obstructing traffic lanes and increasing traffic congestion, and leading city officials to argue that self driving vehicles are taking a toll on city services\cite{lu_2023_lost}. 

Described in this section and summarized in Table~\ref{AV Incident Types}, we present various cases where premature or poorly stopping lead to problems which could have been avoided if a more human interactive framework were used. These cases illustrate that, though stopping ought to be a part of a danger response system, it can also lead to unforeseen consequences. For example, a pedestrian had been run over and pinned under a tire by an AV after assessing its surrounding environment, and it deemed it to be an unsafe situation and came to a sudden stop, which in the end caused more harm\cite{theassociatedpress_2023_cruise}. 

Reports in California have described situations where AVs obstruct the paths of first responders for a few minutes and thereby cause significant damage by delaying their response \cite{lu_2023_lost}. In another case, where a Fire Department was responding to an emergency when an autonomous service robotics cart hesitated when crossing the road, which ultimately caused the first responders to lose precious time\cite{dallow_2025_robot}, reflecting a failure of the autonomous system (in this case, a non-AV robot using transportation infrastructure) to identify first responders during an emergency. There have been multiple cases where a vehicle has failed to respond to authority figures, endangering human drivers by driving through busy intersections \cite{white_2023_san, lu_2023_lost, stone_2023_2}. These scenarios show how rigid safety responses in AV systems do not take social understanding into account when deciding what is safe. While their intended risk goes down by keeping the platform safe and slowing down, these systems may not adequately recognize or prioritize social urgency in planning their behavior. 

In another class of incident, an autonomous vehicle relied on the MRC framework to assess and respond to extreme weather conditions. In doing so, multiple AVs stopped in suboptimal roadway positions, leading to a situation in which a police officer was forced to intervene and stop the self-driving vehicle because it was obstructing traffic due degradation performance. This obstruction limited the forward visibility of human drivers, reducing their ability to anticipate hazards and make timely decisions\cite{kxan}. Similarly, in a power outage, AVs abruptly stopped in place and obstructed traffic, forcing human drivers to swerve around the stationary vehicles \cite{kafton_2022_video}\cite{ap_2023_waymo_outage}. These incidents illustrate the ways in which AVs' planning or fallback frameworks are capable of prioritizing the internal risk for the vehicle, but neglect how humans naturally adjust to one another on the road while operating alongside human drivers or in novel edge conditions \cite{partnersforautomatedvehicleeducationpave_2024_pave} \cite{nicholson_2023_cpuc}. It further reveals how integrating human knowledge and planning capabilities can help find an optimal and safe spot to pull over or adjust behavior depending on where human drivers are present.

We propose that responding to these long-tail challenges requires incorporation of human interaction in AVs, especially in socially-complex environments. Existing MRC frameworks lack understanding of human cues, such as gestures, instructions from authority figures, and implicit coordination with other human drivers, amplifying disruptions to urban environments rather than improving overall safety. Assisted guidance can allow a human-in-the-loop to influence planning of an AV during an incident rather than relying on rigid but underdetermined safety rules, where the MRC framework cannot fully gauge the safety for the rider and road users. 

\section{Enabling Technologies \& Research Directions}
\subsection{Human-Interactive Perception}
\subsubsection{Authority and Instructional Gesture Recognition in Dynamic Road Environments}

Incidents such as AVs driving through and blocking emergency scenes and failing to comply with police gestures demonstrate that AVs currently struggle with recognizing human authority and multimodal instructions \cite{nicholson_2023_cpuc} \cite{californiastateassembly_2024_assembly} \cite{kafton_2022_video}. This highlights a human-interactive perception gap where AVs are unable to translate human instructions into actionable understanding. 

Consequently, recent research has used Convolutional Neural Networks (CNNs) to learn to recognize authorized traffic controller gestures. Such a system was successful in comprehending the gesture almost 97\% of the time, taking approximately 0.625 seconds to make its decision \cite{mishra_2021_authorized}. Certain commercially-deployed robotaxis now know how to respond to traffic police gestures, according to an independent review of first responder protocols conducted by Tüv Süd \cite{hawkins_2024_waymos}, and such capabilities must be adopted in all AVs as emphasized in policy discussions and safety hearings \cite{nicholson_2023_cpuc}. However, these approaches come with limitations. The gesture-recognition methods are evaluated under controlled conditions, where gesture vocabulary is limited to a certain standard \cite{mishra_2021_authorized}. Additionally, it is unclear how robust these methods would be under occlusion, smoke and adverse weather, which are common conditions in emergency situations \cite{rasouli_2022_autonomous}. The use of Vision-Language Models (VLMs) improve contextual reasoning, but are still very susceptible to domain shift and ambiguity; current research finds that VLMs struggle to interpret a pedestrian’s dynamic gestures and reason accurately in context of the scene \cite{bossen2025can}. Lastly, AVs inconsistently model how ambiguous authority recognition should influence downstream decisions. Thus, it is necessary that authority uncertainty in perception outputs is informed in the planning modules of AVs \cite{lin_2018_the} \cite{arnez_2020_a}.

\subsubsection{Work-zone and temporary closure perception}

From the incidents summarized in Section III, AVs are susceptible to abruptly stopping or obstructive behaviors when entering work zones or temporary closures \cite{nicholson_2023_cpuc} \cite{californiastateassembly_2024_assembly} \cite{partnersforautomatedvehicleeducationpave_2024_pave}. These cases highlight the AV’s failure to perceive traffic control devices and understand how they reconfigure traffic rules. Thus, it is necessary that the AV robustly perceives and comprehends the context of such adverse situations for reliable downstream planning.

At the infrastructure level the U.S. Department of Transportation’s Work Zone Data Exchange (WZDx) initiative makes real-time information about work zones, including lane closures and detours, publicly available---albeit with limited coverage as the system requires implementation and compliance with associated districts' transportation committees \cite{usdepartmentoftransportation_2023_work} \cite{usdepartmentoftransportation_utilizing}. By ingesting authoritative, machine-readable descriptions of temporary roadway changes, AVs can better perceive ambiguous temporary traffic control environments, in addition to benefits in efficient routing. These efforts move perception beyond vision-only inference to include human-informed structured information and intelligent transportation infrastructure support.

\subsection{Human-Interactive Planning}
\subsubsection{Language-grounded scene understanding for interactive commands}

In addition to noncompliance with gestures, incidents in Section III highlight current AVs’ inability to comprehend verbal instructions and translate them to actionable planning, including multiple instances where vehicles parked in the middle of traffic but police officers were unable to get them to move ahead without further remote escalation \cite{nicholson_2023_cpuc} \cite{californiastateassembly_2024_assembly} \cite{partnersforautomatedvehicleeducationpave_2024_pave}. 

As a result, it is vital that AVs incorporate human-interactive planning to make human-informed decisions. One of the ways of doing this is through language-grounded scene understanding so that the vehicle can appropriately respond to interactive commands given by first responders during uncertain scenes. This need led to the development of datasets such as Talk2Car \cite{deruyttere_2019_talk2car} and doScenes \cite{roy_2024_doscenes}, which align linguistic input, phrased as natural-language instructions, with visual and environmental representations. From these datasets, multimodal models have been developed to spatially identify the objects referenced in the human instructions under uncertain situations \cite{deruyttere_2021_giving} and plan trajectories from language instructions \cite{thierryderuyttere_2022_talk2car, martinez2026natural}, making for more robust interpretation of commands like “pull over at the curb”, or “follow behind the red car”, and suggesting that following planning instructions not only requires lexical understanding but also reasoning over scene geometry and object relations and constraints. 

Nevertheless, it is necessary to be cautious when integrating these methods into current AV planning stacks. Firstly, these models were trained and tested on simulated or constrained environments. Real-world roads however contain higher perceptual noise and safety critical consequences. \cite{deruyttere_2019_talk2car} Additionally, language commands can possibly be incomplete or conflicting. Multimodal fusion models can reduce ambiguity but misinterpretation is still a concern in AVs, especially when the scene is uncertain or partially occluded \cite{deruyttere_2021_giving}. AVs must be able to evaluate their confidence in command interpretation as well as predicted trajectory planning before executing human-interactive commands. 

Existing gesture-recognition systems already demonstrate promising performance under controlled conditions \cite{mishra_2021_authorized}, while commercially deployed robotaxis have begun incorporating traffic-officer interaction protocols \cite{hawkins_2024_waymos}. A practical next step is extending these systems beyond gesture classification and toward uncertainty-aware authority recognition, where perception outputs include confidence estimates regarding both gesture interpretation and source legitimacy. Such confidence estimates could then be incorporated into downstream planning modules, particularly during emergency scenes where AVs must decide whether temporary human direction should override default driving policies.

\subsubsection{Curbside and accessibility-aware pick-up and drop-off planning}

AVs fail to consider the accessibility concerns at PUDO locations as discussed through the incidents in Section II and III. Curbs are regulated and designed to ensure that they are accessible to everyone, thus, stopping in unsuitable curb zones or blocking bike lanes reflect a planning gap \cite{nicholson_2023_cpuc} \cite{mkaufman_2022_autonomous} \cite{dredf_2023_autonomous}. The Open Mobility Foundation’s Curb Data Specification (CDS) provides a standardized digital framework for representing curb regulations, accessibility and usage information, and loading zone locations \cite{openmobilityfoundation_2022_about}. Additionally, the city of San Francisco handles the Curb Ramp Information System (CRIS), which provides and regularly updates geospatial information about accessible curb ramps \cite{sanfranciscopublicworks_2026_city}. Through machine-readable curb data and locations, AVs have access to information which assists in planning trajectories that comply with the accessibility needs of the passenger. However, accessible PUDO is not a static event, but also a negotiation involving spatial precision and multimodal signaling between the AV and passenger, requiring fine-grained planning \cite{urbanroboticsfoundation_2025_pickup}. For instance, a recent study looks at optimization of PUDO locations based on curb space demand measured through Network kernel density estimation, allowing AVs to better plan PUDO to comply with limited curb space in busy streets \cite{wang_2025_networkbased}. Another recent survey paper demonstrates how human-machine interaction through preset user preferences can make accessibility-aware decisions across all parts of the ride-share experience, from optimized PUDO, to accessible UI interaction, and even adjustment to speed limits and driving style \cite{ashishbastola_2025_driving}. Put together, these technologies leverage infrastructure perception to enforce accessibility-aware planning. Besides integration of human instruction for navigation adjustment, additional limitations include geographical incompletness of curb datasets, which will require efforts and compliance with transportation committees around the globe, and lack of real-time record of temporary obstructions or closures, as updates currently happen once every quarter \cite{openmobilityfoundation_2022_about} \cite{sanfranciscopublicworks_2026_city}. 

Recent language-grounded planning systems demonstrate the ability to associate natural-language instructions with scene geometry and trajectory generation \cite{deruyttere_2019_talk2car}. However, current systems largely assume that user instructions are valid and unambiguous. A necessary extension for real-world deployment is integrating confidence estimation and safety validation \cite{greer2021trajectory} into the planning pipeline before execution of language-conditioned actions. Future AV systems may need to evaluate whether instructions such as “pull over here” are feasible, lawful, and safe under current environmental conditions before generating a final trajectory.

\subsection{Human-Interactive Control: fleet-managed remote assistance and teleoperation}

Many of the incidents discussed in Section III suggested that intervention for failing AVs required intervention from first responders, either to get remote assistance or manually disable the self-driving capabilities of the vehicle. These complex procedures require time and resources from first responders, who have greater priorities \cite{nicholson_2023_cpuc} \cite{californiastateassembly_2024_assembly}. Technological advancements in remote assistance and teleoperation could help reduce this bottleneck.

Remote assistance, defined by the AVSC Best Practice on ADS Remote Assistance, is when a remote human operator is providing guidance to the AV when it enters an MRC, rather than directly controlling it \cite{automatedvehiclesafetyconsortium_2023_avsc}. A commercially-deployed robotaxi company stated to U.S. lawmakers that their remote operators only advise the AV system when requested, and do not “control, steer, or drive”. This means that the AV is still the primary decision-maker, but is helped with additional context to reduce ambiguity and uncertainty. Certain AV companies are currently working on improving the accuracy of uncertainty faced by the vehicles, and has a team of 70 remote operators at any time to deal with failures \cite{shepardson_2026_waymo}. 

In contrast, teleoperation refers to situations when the remote human operator has a more direct role in driving the AV. There is not a clear taxonomy on the different approaches of remote operation, thus Bogdoll et al. outlined specifications and requirements in a survey paper to help the scientific community and policymakers make better decisions about the usage of one over the other \cite{bogdoll_2022_taxonomy}, especially because teleoperation causes concerns over risks from network latency and connectivity \cite{kamtam_2024_network}. Similarly, a study by Lu et al. theorizes a teleoperation framework that better monitors and partially controls AVs to allow for more control, yet ensures that autonomy is preserved to govern safe performance \cite{lu2022teleoperation}. Real-world tele-driving deployments suggest that a complete remote-driving paradigm is achievable; through the use of a 360-degree camera view, wheels, pedals, and even a headset to hear sounds outside the vehicle, a remote operator is able to drive the vehicle. Multi-network connectivity strategies, as well as other policies, align the technology with the highest safety and regulatory standards \cite{gillespie_2025_how}.

Open challenges still remain with these developments. “Guidance” in remote assistance must be formalized in a way that AV systems can interpret such guidance safely without ambiguity. It is also necessary to continue reducing latency experienced in teleoperations, and ensure that fall-back mechanisms are robust, again imposing the need for human-interactive autonomous systems beyond remote control. 

\section{Concluding Remarks}

This paper examines a growing gap between the traditional safety philosophy of autonomous vehicles and the realities of operating in complex, human-governed, dynamic, social urban environments. While the MRC framework is effective for mitigating immediate collision risk, evidence from publicly documented incidents shows that stopping alone does not always lead to safe outcomes. In many cases, rigid fallback behaviors can introduce new hazards such as obstructing traffic, interfering with emergency response operations, or failing to accommodate accessibility needs. These incidents highlight that safety in real-world roadway ecosystems requires more than collision avoidance; it requires cooperative and context-aware interaction with humans who actively regulate traffic and infrastructure.

Through the analysis of documented incidents, this work presents a case study of failures that can be traced to limitations in perception, planning, and control when systems encounter socially dynamic scenarios. The cases reviewed demonstrate that current AV architectures are often optimized for rule-based traffic environments but struggle when authority figures, temporary roadway changes, or accessibility-related interactions require flexible interpretation and response. These findings emphasize that real-world deployment demands systems capable of understanding human intent, recognizing authority signals, and adapting to situational context beyond predefined traffic rules.

In addition, these cases show current difficulties plaguing AV incident reporting. Currently, incidents are reported inconsistently, and different private and public organizations have separate reporting systems, leading to poor understanding of the complete AV incident landscape \cite{californiastateassembly_2024_assembly}. The incidents compiled into Table~\ref{AV Incident Types} only represent a small sample of the premature stopping issues which hinder AVs. Without a centralized reporting system it is difficult to fully identify the key weaknesses of AVs, making it difficult to address challenges. Though they are not comprehensive, the samples included in Table 1 aim to capture a representative sample of the human interactivity problems which AVs face.

To address limitations in autonomous systems, this paper discusses emerging research directions in human-interactive perception, language-grounded planning, accessibility-aware decision making, and assisted control mechanisms such as remote guidance and teleoperation. Together, these approaches suggest a shift in AV design philosophy: from purely autonomous operation toward cooperative autonomy that incorporates human input, contextual awareness, and infrastructure-informed decision making. However, significant challenges remain, including robustness under uncertainty, standardization of interaction protocols, scalability of remote assistance, and the availability of reliable data describing dynamic road conditions. 

Ultimately, enabling safe and reliable AV deployment in urban environments will require integrating human-interactive capabilities directly into the perception, planning, and control stack. Future work should focus on bridging the gap between technical autonomy and real-world social driving dynamics through improved datasets, evaluation frameworks, and regulatory coordination. By advancing these capabilities, autonomous vehicles can move beyond passive fallback strategies and become cooperative participants in the complex human systems that govern modern roadways.

\bibliographystyle{IEEEtran}
\bibliography{refs}

\begin{thebibliography}{10}
\providecommand{\url}[1]{#1}
\csname url@samestyle\endcsname
\providecommand{\newblock}{\relax}
\providecommand{\bibinfo}[2]{#2}
\providecommand{\BIBentrySTDinterwordspacing}{\spaceskip=0pt\relax}
\providecommand{\BIBentryALTinterwordstretchfactor}{4}
\providecommand{\BIBentryALTinterwordspacing}{\spaceskip=\fontdimen2\font plus
\BIBentryALTinterwordstretchfactor\fontdimen3\font minus \fontdimen4\font\relax}
\providecommand{\BIBforeignlanguage}[2]{{%
\expandafter\ifx\csname l@#1\endcsname\relax
\typeout{** WARNING: IEEEtran.bst: No hyphenation pattern has been}%
\typeout{** loaded for the language `#1'. Using the pattern for}%
\typeout{** the default language instead.}%
\else
\language=\csname l@#1\endcsname
\fi
#2}}
\providecommand{\BIBdecl}{\relax}
\BIBdecl

\bibitem{lee_2024_waymos}
\BIBentryALTinterwordspacing
T.~B. Lee, ``Waymo’s investments in san francisco may be paying off,'' Understanding AI, 05 2024. [Online]. Available: \url{https://www.understandingai.org/p/waymos-investments-in-san-francisco}
\BIBentrySTDinterwordspacing

\bibitem{partnersforautomatedvehicleeducationpave_2024_pave}
\BIBentryALTinterwordspacing
Z.~R. Adam~Campbell, Bart~Teeter, ``Pave virtual panel "open roads: Av operations in texas - part 2",'' YouTube, March 2024. [Online]. Available: \url{https://www.youtube.com/watch?v=-WfXGA8mQDU}
\BIBentrySTDinterwordspacing

\bibitem{dagostino_2024_experiences}
\BIBentryALTinterwordspacing
M.~C. D'Agostino, C.~E. Michael, and P.~S. Venkataram, ``Experiences with autonomous vehicle in us cities,'' 2024. [Online]. Available: \url{https://www.brookings.edu/articles/how-autonomous-vehicles-could-change-cities/}
\BIBentrySTDinterwordspacing

\bibitem{zhou_2023_sf}
\BIBentryALTinterwordspacing
Y.~Zhou, ``Sf cops, firefighters vent before big vote on driverless car expansion,'' Mission Local, 08 2023. [Online]. Available: \url{https://missionlocal.org/2023/08/sf-cops-firefighters-vent-prior-to-big-vote-on-driverless-car-expansion/}
\BIBentrySTDinterwordspacing

\bibitem{nicholson_2023_cpuc}
\BIBentryALTinterwordspacing
J.~Nicholson, D.~Luttropp, N.~Jones, and J.~Friedlander, ``Cpuc status conference: Safety issues regarding driverless av interactions with first responders,'' 08 2023. [Online]. Available: \url{https://www.sfmta.com/sites/default/files/reports-and-documents/2023/08/2023.08.07_cpuc_status_conference_8.7.2023_final.pdf}
\BIBentrySTDinterwordspacing

\bibitem{tumlin_2023_san}
\BIBentryALTinterwordspacing
J.~P. Tumlin, T.~Chang, and R.~Hillis, ``{San Francisco}'s application to rehear resolution {TL-19145} approving authorization for {Cruise LLC}'s expanded service in autonomous vehicle passenger service phase {I} driverless deployment program,'' City and County of San Francisco, Tech. Rep., Sep. 2023, accessed: Mar. 1, 2026. [Online]. Available: \url{https://docs.cpuc.ca.gov/PublishedDocs/Efile/G000/M520/K495/520495874.PDF}
\BIBentrySTDinterwordspacing

\bibitem{californiapublicutilitiescommission_2023_re}
\BIBentryALTinterwordspacing
N.~B. Jeffrey~Tumlin, Tilly~Chang, ``Re: Protest of cruise llc tier 2 advice letter (0002),'' Jan 2023. [Online]. Available: \url{https://www.sfmta.com/sites/default/files/reports-and-documents/2023/01/2023.01.25_ccsf_23.0125_cpuc_cruise_tier_2_advice_letter_protest_002.pdf}
\BIBentrySTDinterwordspacing

\bibitem{white_2023_san}
\BIBentryALTinterwordspacing
J.~B. White, ``San francisco moves to block robotaxi expansion,'' Politico, 08 2023. [Online]. Available: \url{https://www.politico.com/news/2023/08/17/san-francisco-block-robotaxi-expansion-00111657}
\BIBentrySTDinterwordspacing

\bibitem{onroadautomateddrivingoradcommittee_2021_j3016_202104}
\BIBentryALTinterwordspacing
{On-Road Automated Driving (ORAD) Committee}, ``J3016\_202104 - taxonomy and definitions for terms related to driving automation systems for on-road motor vehicles,'' 04 2021. [Online]. Available: \url{https://www.sae.org/standards/j3016_202104-taxonomy-definitions-terms-related-driving-automation-systems-road-motor-vehicles}
\BIBentrySTDinterwordspacing

\bibitem{automateddrivingbehaviorsconsortium_2019_minimal}
\BIBentryALTinterwordspacing
{Automated Driving Behaviors Consortium}, ``Minimal risk condition behaviors based on environmental and vehicle conditions,'' 06 2019. [Online]. Available: \url{https://pronto-core-cdn.prontomarketing.com/2/wp-content/uploads/sites/2896/2019/07/Minimal-Risk-Condition-Behaviors-June-2019-FINAL.pdf}
\BIBentrySTDinterwordspacing

\bibitem{ding_2025_waymos}
\BIBentryALTinterwordspacing
J.~Ding and M.~Liedtke, ``Waymos in san francisco caused chaos during mass power outage,'' AP News, 12 2025. [Online]. Available: \url{https://apnews.com/article/waymo-cars-san-francisco-power-outage-traffic-81e6a00aa2be6b804fe0bdfbcf07401f}
\BIBentrySTDinterwordspacing

\bibitem{lu_2023_lost}
\BIBentryALTinterwordspacing
Y.~Lu, ``‘lost time for no reason’: How driverless taxis are stressing cities,'' 11 2023. [Online]. Available: \url{https://www.nytimes.com/2023/11/20/technology/driverless-taxis-cars-cities.html}
\BIBentrySTDinterwordspacing

\bibitem{cobler_2026_waymo}
\BIBentryALTinterwordspacing
N.~Cobler, ``Waymo robotaxi blocks ems responding to austin mass shooting,'' Axios, 03 2026. [Online]. Available: \url{https://www.axios.com/local/austin/2026/03/02/waymo-vehicle-blocks-ems-austin-mass-shooting}
\BIBentrySTDinterwordspacing

\bibitem{californiastateassembly_2024_assembly}
\BIBentryALTinterwordspacing
{California State Assembly and CalMatters Digital Democracy}, ``Assembly standing committee on transportation,'' 04 2024. [Online]. Available: \url{https://calmatters.digitaldemocracy.org/hearings/257824#t=1549&f=990acfa7cc45c1e0dec9c004c588c025}
\BIBentrySTDinterwordspacing

\bibitem{mkaufman_2022_autonomous}
\BIBentryALTinterwordspacing
S.~M.~Kaufman, J.~Y.~J.~Chow, B.~Liu, A.~Yamron, and M.~Geck, ``Autonomous vehicle good citizenry standard,'' 07 2022. [Online]. Available: \url{https://rosap.ntl.bts.gov/view/dot/67860/dot_67860_DS1.pdf}
\BIBentrySTDinterwordspacing

\bibitem{yousfi_2025_automated}
E.~Yousfi, T.~Jacquet, and N.~M{\'e}tayer, ``Automated vehicles and people living with a disability: Opportunities, challenges, and future directions for sustainable mobility,'' \emph{Sustainability}, vol.~17, no.~13, p. 5941, 2025.

\bibitem{dredf_2023_autonomous}
\BIBentryALTinterwordspacing
DREDF, ``Autonomous vehicle safety and accessibility for people with disabilities - dredf,'' DREDF, 03 2023. [Online]. Available: \url{https://dredf.org/autonomous-vehicle-safety-and-accessibility-for-people-with-disabilities/}
\BIBentrySTDinterwordspacing

\bibitem{lin_2018_the}
S.-C. Lin, Y.~Zhang, C.-H. Hsu, M.~Skach, M.~E. Haque, L.~Tang, and J.~Mars, ``The architectural implications of autonomous driving: Constraints and acceleration,'' in \emph{Conference on architectural support for programming languages and operating systems}, 2018, pp. 751--766.

\bibitem{nebot2026eraendtoendautonomytransitioning}
\BIBentryALTinterwordspacing
E.~Nebot and J.~S.~B. Perez, ``The era of end-to-end autonomy: Transitioning from rule-based driving to large driving models,'' 2026. [Online]. Available: \url{https://arxiv.org/abs/2603.16050}
\BIBentrySTDinterwordspacing

\bibitem{hu_2022_planningoriented}
Y.~Hu, J.~Yang, L.~Chen, K.~Li, C.~Sima, X.~Zhu, S.~Chai, S.~Du, T.~Lin, W.~Wang \emph{et~al.}, ``Planning-oriented autonomous driving,'' in \emph{Proceedings of the IEEE/CVF conference on computer vision and pattern recognition}, 2023, pp. 17\,853--17\,862.

\bibitem{greer2025language}
R.~Greer, B.~Antoniussen, A.~M{\o}gelmose, and M.~Trivedi, ``Language-driven active learning for diverse open-set 3d object detection,'' in \emph{Proceedings of the Winter Conference on Applications of Computer Vision}, 2025, pp. 980--988.

\bibitem{advocatesforhighwayandautosafety_2020_autonomous}
\BIBentryALTinterwordspacing
{Advocates for Highway and Auto Safety}, ``Autonomous vehicle (av) tenets,'' 11 2020. [Online]. Available: \url{https://saferoads.org/wp-content/uploads/2020/11/AV-Tenets-11-24-20-1.pdf}
\BIBentrySTDinterwordspacing

\bibitem{cano_2024_exclusive}
\BIBentryALTinterwordspacing
R.~Cano, ``Exclusive: Driverless robotaxis are causing less mayhem on s.f. streets. city officials explain why,'' \emph{San Francisco Chronicle}, 02 2024. [Online]. Available: \url{https://www.sfchronicle.com/sf/article/driverless-robotaxis-incidents-decrease-18672791.php}
\BIBentrySTDinterwordspacing

\bibitem{arnez_2020_a}
F.~Arnez, H.~Espinoza, A.~Radermacher, and F.~Terrier, ``A comparison of uncertainty estimation approaches in deep learning components for autonomous vehicle applications,'' \emph{arXiv preprint arXiv:2006.15172}, 2020.

\bibitem{saeinternational_2024_195}
\BIBentryALTinterwordspacing
{SAE International}, ``195. partnering with law enforcement for safer avs,'' YouTube, 08 2024. [Online]. Available: \url{https://www.youtube.com/watch?v=TvhSmCwtmJw}
\BIBentrySTDinterwordspacing

\bibitem{roy_2024_doscenes}
P.~Roy, S.~Perisetla, S.~Shriram, H.~Krishnaswamy, A.~Keskar, and R.~Greer, ``doscenes: An autonomous driving dataset with natural language instruction for human interaction and vision-language navigation,'' \emph{IEEE Intelligent Transportation Systems Conference}, 2025.

\bibitem{chang_2024_it}
X.~Chang, Z.~Chen, X.~Dong, Y.~Cai, T.~Yan, H.~Cai, Z.~Zhou, G.~Zhou, and J.~Gong, ``"it must be gesturing towards me": Gesture-based interaction between autonomous vehicles and pedestrians,'' in \emph{Proceedings of the 2024 CHI Conference on Human Factors in Computing Systems}, 2024, pp. 1--25.

\bibitem{agent2025p3474}
V.~A. V.~S. Committee, ``P3474/d2, apr 2025 - ieee draft standard for human intentions and artificial intelligence alignment in autonomous driving agent,'' \emph{IEEE P3474/D2}, 2025.

\bibitem{rasouli_2022_autonomous}
A.~Rasouli and J.~K. Tsotsos, ``Autonomous vehicles that interact with pedestrians: A survey of theory and practice,'' \emph{IEEE transactions on intelligent transportation systems}, vol.~21, no.~3, pp. 900--918, 2019.

\bibitem{shriram2025towards}
S.~Shriram, S.~Perisetla, A.~Keskar, H.~Krishnaswamy, T.~E.~W. Bossen, A.~M{\o}gelmose, and R.~Greer, ``Towards a multi-agent vision-language system for zero-shot novel hazardous object detection for autonomous driving safety,'' in \emph{2025 IEEE 21st International Conference on Automation Science and Engineering (CASE)}.\hskip 1em plus 0.5em minus 0.4em\relax IEEE, 2025, pp. 1511--1518.

\bibitem{chen2025robust}
Y.~Chen and R.~Greer, ``Robust scenario mining assisted by multimodal semantics,'' in \emph{Proceedings of the IEEE/CVF International Conference on Computer Vision}, 2025, pp. 1792--1801.

\bibitem{dvilasobern_2025_a}
S.~D{\'a}vila-Sober{\'o}n, A.~Morales-D{\'\i}az, and M.~Castel{\'a}n, ``A novel image dataset for detecting and classifying mobility aid users,'' \emph{Expert Systems with Applications}, vol. 293, p. 128697, 2025.

\bibitem{martnezchao_2024_urban}
T.~E. Martínez-Chao, A.~Menéndez-Díaz, S.~García-Cortés, and P.~D’Agostino, ``Urban pedestrian routes’ accessibility assessment using geographic information system processing and deep learning-based object detection,'' \emph{Sensors}, vol.~24, pp. 3667--3667, 06 2024.

\bibitem{omeara_2022_rampnet}
J.~S. O'Meara, J.~Hwang, Z.~Wang, M.~Saugstad, and J.~E. Froehlich, ``Rampnet: A two-stage pipeline for bootstrapping curb ramp detection in streetscape images from open government metadata,'' in \emph{Proceedings of the IEEE/CVF International Conference on Computer Vision}, 2025, pp. 6656--6665.

\bibitem{kafton_2022_video}
\BIBentryALTinterwordspacing
C.~Kafton, ``Video: Driverless car pulled over by san francisco police,'' KTVU FOX 2 San Francisco, 04 2022. [Online]. Available: \url{https://www.ktvu.com/news/video-driverless-car-pulled-over-by-san-francisco-police}
\BIBentrySTDinterwordspacing

\bibitem{dallow_2025_robot}
\BIBentryALTinterwordspacing
L.~Dallow, ``Robot delivery cart blocks fire truck in hollywood intersection, video shows,'' KTLA, 09 2025. [Online]. Available: \url{https://ktla.com/news/local-news/robot-delivery-cart-blocks-fire-truck-in-hollywood-intersection-video-shows/}
\BIBentrySTDinterwordspacing

\bibitem{kxan}
\BIBentryALTinterwordspacing
E.~Pauda, ``Apd officers move waymo vehicles pulled over during severe weather,'' kxan, 2025. [Online]. Available: \url{https://www.kxan.com/news/local/austin/apd-officers-successfully-and-safely-move-waymo-vehicles-blocking-traffic/}
\BIBentrySTDinterwordspacing

\bibitem{theassociatedpress_2023_cruise}
\BIBentryALTinterwordspacing
T.~A. Press and O.~Palma, ``Cruise suspends driverless robotaxi service nationwide,'' 10 2023. [Online]. Available: \url{https://www.kqed.org/news/11965752/cruise-suspends-driverless-robotaxi-service-nationwide}
\BIBentrySTDinterwordspacing

\bibitem{stone_2023_2}
\BIBentryALTinterwordspacing
J.~Stone, ``2 waymo self-driving cars stall at sf pride parade street closures during heavy traffic,'' ABC7 San Francisco, 06 2023. [Online]. Available: \url{https://abc7news.com/post/waymo-stalled-self-driving-car-sf-pride-robotaxi/13427435/}
\BIBentrySTDinterwordspacing

\bibitem{ap_2023_waymo_outage}
\BIBentryALTinterwordspacing
{Associated Press}, ``Waymos blocked roads and caused chaos during san francisco power outage,'' \emph{AP News}, 12 2025. [Online]. Available: \url{https://apnews.com/article/waymo-cars-san-francisco-power-outage-traffic-81e6a00aa2be6b804fe0bdfbcf07401f}
\BIBentrySTDinterwordspacing

\bibitem{mishra_2021_authorized}
A.~Mishra, J.~Kim, J.~Cha, D.~Kim, and S.~Kim, ``Authorized traffic controller hand gesture recognition for situation-aware autonomous driving,'' \emph{Sensors}, vol.~21, no.~23, p. 7914, 2021.

\bibitem{hawkins_2024_waymos}
\BIBentryALTinterwordspacing
A.~J. Hawkins, ``Waymo’s robotaxis pass the first responder test,'' The Verge, 12 2024. [Online]. Available: \url{https://www.theverge.com/2024/12/13/24319860/waymo-robotaxi-first-responder-tuv-sud-analysis}
\BIBentrySTDinterwordspacing

\bibitem{bossen2025can}
T.~E. Bossen, A.~M{\o}gelmose, and R.~Greer, ``Can vision-language models understand and interpret dynamic gestures from pedestrians? pilot datasets and exploration towards instructive nonverbal commands for cooperative autonomous vehicles,'' in \emph{Proceedings of the Computer Vision and Pattern Recognition Conference}, 2025, pp. 4779--4788.

\bibitem{usdepartmentoftransportation_2023_work}
\BIBentryALTinterwordspacing
{U.S. Department of Transportation}, ``Work zone data exchange (wzdx) | us department of transportation,'' Transportation.gov, 2023. [Online]. Available: \url{https://www.transportation.gov/av/data/wzdx}
\BIBentrySTDinterwordspacing

\bibitem{usdepartmentoftransportation_utilizing}
\BIBentryALTinterwordspacing
U.~D. of~Transportation, ``Utilizing work zone event data in connected and smarter work zone applications-maricopa county, arizona regional work zone management and operations overview.'' [Online]. Available: \url{https://ops.fhwa.dot.gov/publications/fhwahop20022/fhwahop20022.pdf}
\BIBentrySTDinterwordspacing

\bibitem{deruyttere_2019_talk2car}
T.~Deruyttere, S.~Vandenhende, D.~Grujicic, L.~Van~Gool, and M.~F. Moens, ``Talk2car: Taking control of your self-driving car,'' in \emph{Conference on empirical methods in natural language processing and joint conference on natural language processing}, 2019, pp. 2088--2098.

\bibitem{deruyttere_2021_giving}
T.~Deruyttere, V.~Milewski, and M.-F. Moens, ``Giving commands to a self-driving car: How to deal with uncertain situations?'' \emph{Engineering Applications of Artificial Intelligence}, vol. 103, p. 104257, 08 2021.

\bibitem{thierryderuyttere_2022_talk2car}
T.~Deruyttere, D.~Grujicic, M.~B. Blaschko, and M.-F. Moens, ``Talk2car: Predicting physical trajectories for natural language commands,'' \emph{IEEE Access}, vol.~10, pp. 123\,809--123\,834, 01 2022.

\bibitem{martinez2026natural}
A.~Martinez-Sanchez, P.~Roy, and R.~Greer, ``Natural language instructions for scene-responsive human-in-the-loop motion planning in autonomous driving using vision-language-action models,'' \emph{IEEE Intelligent Vehicles Symposium}, 2026.

\bibitem{openmobilityfoundation_2022_about}
\BIBentryALTinterwordspacing
{Open Mobility Foundation}, ``About {CDS} | open mobility foundation,'' 01 2022. [Online]. Available: \url{https://www.openmobilityfoundation.org/about-cds/}
\BIBentrySTDinterwordspacing

\bibitem{sanfranciscopublicworks_2026_city}
\BIBentryALTinterwordspacing
{San Francisco Public Works}, ``City of san francisco - curb ramps,'' Data.gov, 2026. [Online]. Available: \url{https://catalog-beta.data.gov/dataset/curb-ramps}
\BIBentrySTDinterwordspacing

\bibitem{urbanroboticsfoundation_2025_pickup}
\BIBentryALTinterwordspacing
{Urban Robotics Foundation}, ``Pick-up and drop-off at the curb,'' Urban Robotics Fdn., 05 2025. [Online]. Available: \url{https://www.urbanroboticsfoundation.org/post/pick-up-and-drop-off-at-the-curb}
\BIBentrySTDinterwordspacing

\bibitem{wang_2025_networkbased}
S.~Wang, F.~Ding, Y.~Wang, A.~Gar~on Yeh, and G.~Huang, ``Network-based pick-up and drop-off location optimization for shared autonomous vehicles,'' \emph{Urban Informatics}, vol.~4, 05 2025.

\bibitem{ashishbastola_2025_driving}
A.~Bastola, H.~Wang, P.~H.~B. Sayed, J.~Brinkley, A.~J. Moshayedi, and A.~Razi, ``Driving towards inclusion: A systematic review of ai-powered accessibility enhancements for people with disability in autonomous vehicles,'' in \emph{IEEE Access}, 2025.

\bibitem{greer2021trajectory}
R.~Greer, N.~Deo, and M.~Trivedi, ``Trajectory prediction in autonomous driving with a lane heading auxiliary loss,'' \emph{IEEE Robotics and Automation Letters}, vol.~6, no.~3, pp. 4907--4914, 2021.

\bibitem{automatedvehiclesafetyconsortium_2023_avsc}
\BIBentryALTinterwordspacing
{Automated Vehicle Safety Consortium}, ``Avsc releases best practice for automated driving system remote assistance use case,'' Sae-itc.com, 11 2023. [Online]. Available: \url{https://avsc.sae-itc.com/news/avsc-releases-best-practice-automated-driving-system-remote-assistance-use-case}
\BIBentrySTDinterwordspacing

\bibitem{shepardson_2026_waymo}
\BIBentryALTinterwordspacing
D.~Shepardson, ``Waymo defends use of remote assistance workers in robotaxi operations,'' \emph{Reuters}, 02 2026. [Online]. Available: \url{https://www.reuters.com/technology/waymo-defends-use-remote-assistance-workers-robotaxi-operations-2026-02-17/}
\BIBentrySTDinterwordspacing

\bibitem{bogdoll_2022_taxonomy}
D.~Bogdoll, S.~Orf, L.~Töttel, and J.~M. Zöllner, ``Taxonomy and survey on remote human input systems for driving automation systems,'' in \emph{Proceedings of FICC}, 2022.

\bibitem{kamtam_2024_network}
S.~B. Kamtam, Q.~Lu, F.~Bouali, O.~C.~L. Haas, and S.~Birrell, ``Network latency in teleoperation of connected and autonomous vehicles: A review of trends, challenges, and mitigation strategies,'' \emph{Sensors}, vol.~24, pp. 3957--3957, 06 2024.

\bibitem{lu2022teleoperation}
S.~Lu, R.~Zhong, and W.~Shi, ``Teleoperation technologies for enhancing connected and autonomous vehicles,'' in \emph{2022 IEEE 19th International Conference on Mobile Ad Hoc and Smart Systems (MASS)}, 2022.

\bibitem{gillespie_2025_how}
\BIBentryALTinterwordspacing
D.~Gillespie, ``How we make remote driving possible - vay,'' Vay, 12 2025. [Online]. Available: \url{https://vay.io/how-we-make-remote-driving-possible/}
\BIBentrySTDinterwordspacing

\end{thebibliography}

\end{document}